\documentclass[10pt,twocolumn,letterpaper]{article}

\usepackage{iccv}
\usepackage{times}
\usepackage{epsfig}
\usepackage{graphicx}
\usepackage{amsmath}
\usepackage{amssymb}
\usepackage{subfigure}

\usepackage[pagebackref=true,breaklinks=true,letterpaper=true,colorlinks,bookmarks=false]{hyperref}

\iccvfinalcopy

\def\iccvPaperID{2458} 

\ificcvfinal\fi
\begin{document}

\title{Jointly Attentive Spatial-Temporal Pooling Networks for Video-based\\ Person Re-Identification}

\author{ 
Shuangjie Xu$^{1}$\thanks{indicates equal contributions.}\,\,  \hspace{0.1in} Yu Cheng$^{2*}$  \hspace{0.1in} Kang Gu$^1$   \hspace{0.1in} Yang Yang$^3$ \hspace{0.1in} Shiyu Chang$^4$ \hspace{0.1in} Pan Zhou$^1$\\ \\
$^1$Huazhong University of Science and Technology \hspace{0.2in} $^2$AI Foundations, IBM Research\hspace{0.2in} \\ $^4$Northwestern University \hspace{0.2in} $^3$IBM T.J. Watson Research Center \hspace{0.2in}
}

\maketitle

\begin{abstract}
Person Re-Identification (person re-id) is a crucial task as its applications in visual surveillance and human-computer interaction.  In this work, we present a novel joint Spatial and Temporal Attention Pooling Network (ASTPN) for video-based person re-identification, which enables the feature extractor to be aware of the current input video sequences, in a way that interdependency from the matching items can directly influence the computation of each other's representation. Specifically, the spatial pooling layer is able to select regions from each frame, while the attention temporal pooling performed can select informative frames over the sequence, both pooling guided by the information from distance matching. Experiments are conduced on the iLIDS-VID, PRID-2011 and MARS datasets and the results demonstrate that this approach outperforms existing state-of-art methods. We also analyze how the joint pooling in both dimensions can boost the person re-id performance more effectively than using either of them separately \footnote{the code is available at https://github.com/shuangjiexu/Spatial-Temporal-Pooling-Networks-ReID}. 
\end{abstract}

\section{Introduction}
Person Re-Identification has been viewed as one of the key subproblems of the generic object recognition task. It is also important due to its applications in surveillance, and human-computer interaction communities. Given a query image, the task is to identify a set of matching person images from a pool, usually captured from the same/different cameras, from different viewpoints, at the same/different time points. It is a very challenging task due to the large variations of lighting conditions, viewing angles, body poses and occlusions. 

Methods for re-identification in still images setting have been extensively investigated, including feature representation learning \cite{Kviatovsky,Ma,Liu,mid_level_filter}, distance metric learning \cite{Liao,R_D_C,kernal_metric,G_M_P,Salience_matching,metric_ensemble} and CNN-based schemes \cite{SCNN,RFA,NIPS2016_6367,DBLP:conf/eccv/VariorSLXW16}. Very recently, researchers began to explore solving this problem in video-based setting, which is a more natural way to perform re-identification. The intuition of this kind of methods is that temporal information related to person motion can be captured from video. Moreover, sequences of images provide rich samples of persons' appearances, helping boosting the re-identification performance with more discriminative features. In \cite{C_RNN}, a temporal deep neural network architecture combines optical flow, recurrent layers and mean-pooling and achieves reasonable success. The work in \cite{RFA} exploited a novel recurrent feature aggregation framework, which is capable of learning discriminative sequence level representation from simple frame-wise features. 

\begin{figure}
\centering
\includegraphics[width=8.5cm]{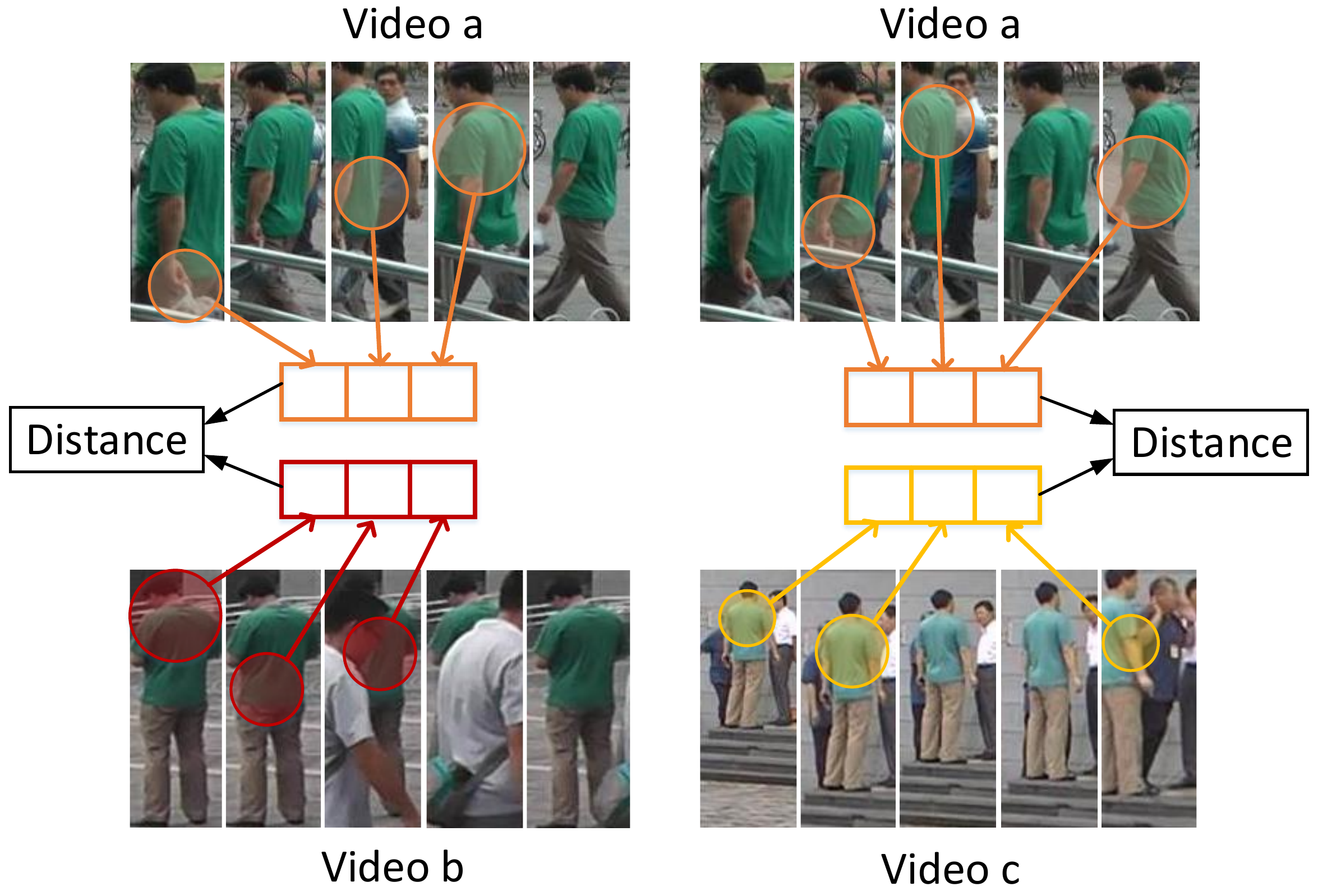}
\caption{Sample video frames from one person captured by three cameras a, b and c, simulating how human compare different video pairs. The regions under the cycles are the parts which visual attentions are drawn on.}
\label{fig:example}
\end{figure}

The main idea of video-based methods is first to extract useful representations from video images with RNN (or CNN-RNN) models. Then exploit a distance function to judge their extent of matching. However, most of these approaches derive each sequence's representation separately, rarely considering the impact of the others, which neglect the mutual influence of the two video sequences in the context of the matching task. Let's think about how human visual processing works when comparing video sequences. For example, the pair-wise case described in Figure \ref{fig:example}, when comparing video frames a with two other \textit{b} and \textit{c} separately, as \textit{b} and \textit{c} are different, it is natural for our brain to draw different focuses on different frames of \textit{a}. On the other hand, the interaction of compared sequences should also have effect on the spatial dimension, which guides human to pay attentions on different regions of the input \textit{a}. This is extremely important for the scenario with large viewpoint changes or fast moving object. The example demonstrates why we should draw different attention when comparing different pairs of video frames. 

Motivated by recent success of attention models \cite{DBLP:journals/corr/BahdanauCB14,DBLP:conf/icml/XuBKCCSZB15,DBLP:journals/tacl/YinSXZ16,Santos2016AttentivePN}, we proposed jointly Attentive Spatial-Temporal Pooling Networks (ASTPN), a powerful mechanism for learning the representation of video sequences by taking into account the interdependence among them. Specifically, ASTPN first learns a similarity measure over the features extracted from recurrent-convolutional networks of the two input items, and uses the similarity scores between the features to compute attention vectors in both spatial (regions in each frame) and temporal (frames over sequences) dimensions. Next, the attention vectors are used to perform pooling. Finally, a Siamese network architecture is deployed over the attention vectors. The proposed architecture can be trained efficiently with the end-to-end training schema.  

We perform extensive experiments on three datasets, iLIDS-VID, PRID-2011 and MARS. The results clearly demonstrate that our proposed method for person re-identification outperforms well established baselines significantly and offers new state-of-the-art performance. The cross dataset test also derives the same conclusion. ASTPN is also a general component that can handle a wide variety of person re-identification tasks.


\begin{figure}[!t]
\centering
\includegraphics[width=8.5cm]{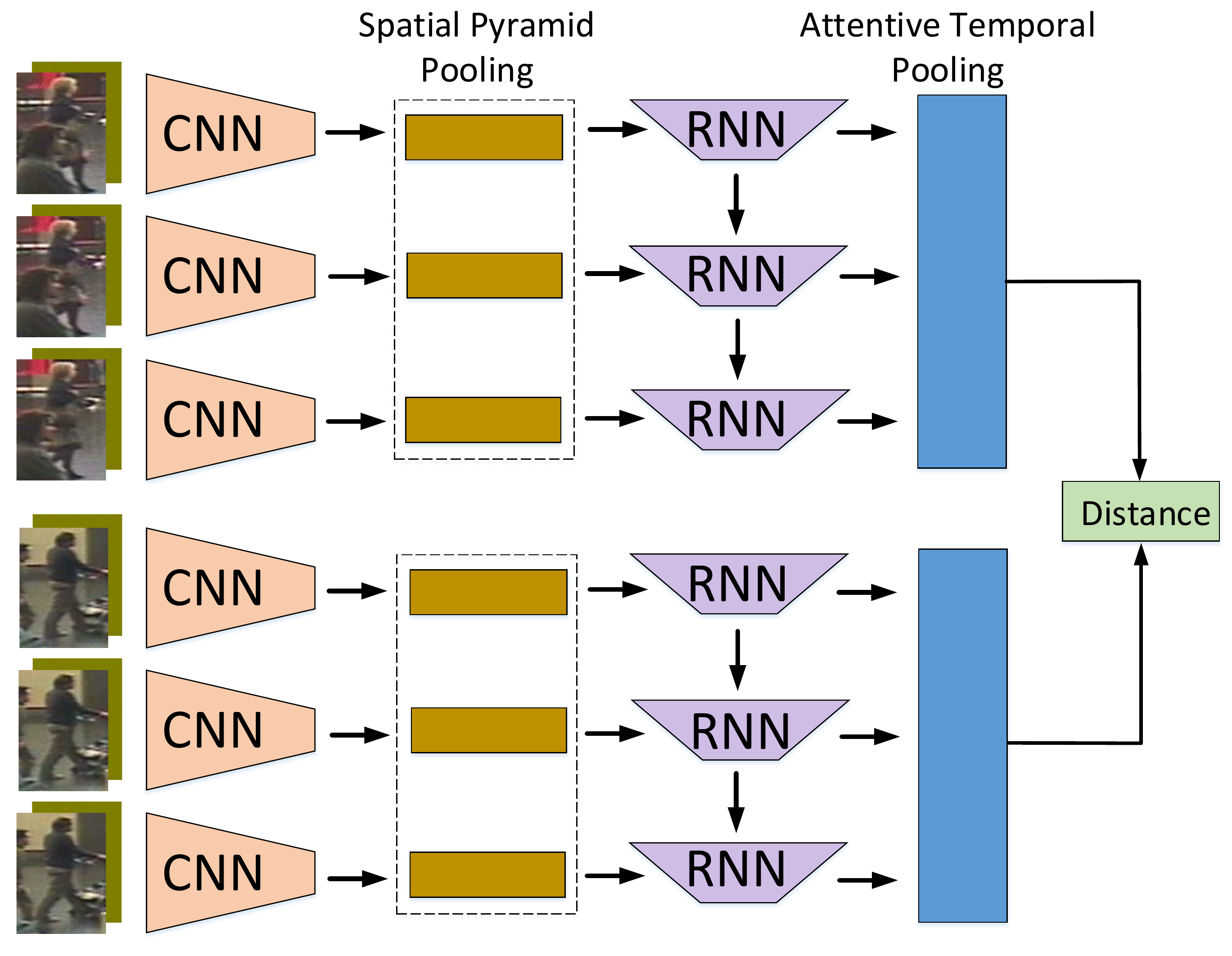}
\caption{Our video-based person re-identification system. We adopt Siamese network architecture for spatial-temporal feature extraction, and jointly attentive spatial-temporal pooling for interdependence information learning.}
\label{fig_main}
\end{figure}

\section{Related Work}
Person re-id, a challenging task which has been explored for several years, still remains to be further focused on to overcome the problems of viewpoint difference, illumination change, occlusions and even similar appearance of different people. A majority of recent works mainly develop their solutions from two aspects: extracting reliable feature representations \cite{Wang,Farenzena,Kviatovsky,Ma,Liu,mid_level_filter} or learning a robust distance metric \cite{Liao,R_D_C,M_metric,kernal_metric,G_M_P,Salience_matching,metric_ensemble,L_A_D_F}. To be specific, features including color histograms\cite{Salience_matching,kernal_metric}, texture histograms \cite{Farenzena}, Local Binary Patterns \cite{kernal_metric} , Color Names \cite{Zheng} and so on are widely utilized for person re-id to address identity information in the existence of challenges like lighting change. In the meantime, metric learning methods such as large margin nearest neighbor (LMNN) \cite{LMNN}, Mahalanobis distance metric (RCA) \cite{M_metric}, Locally Adaptive Decision Function (LADF) \cite{L_A_D_F} and RankSVM \cite{mid_level_filter} have also been applied to person re-id task. Despite the prominent progress in recent years, most of these works are still based on image-to-image level. Video setting is intuitively more close to the practical scenario as video is the first-hand material captured by surveillance camera \cite{Cheng_2014_CVPR,DBLP:conf/icmcs/ChengBFFPZ14}. Besides, temporal information relevant to a person's motion, gait for instance, may help to discriminate similar pedestrians. Moreover, video provides abundant samples of the target for us with the cost of increasing computation.

Gradually, more and more works began to explore video-to-video matching problem in  person re-id. Discriminative Video Ranking model \cite{video_ranking} used discriminative video fragments selection to capture more accurate space-time information, while simultaneously learning a video ranking function for person re-id. Bag-of-words \cite{Zheng} method aimed to encode frame-wise features into a global vector. However, neither of these models could be considered effective for ignoring the rich temporal information contained in the videos. However, video-based person re-id raises new challenges: some inter-class difference of video-based representation can be much more ambiguous compared with the one when using image-based representation, since it's likely that different people could not only have similar appearance but also similar motions, making alignment tough to achieve. Therefore, space-time information must be fully utilized to solve those extra problems. Besides, a top-push distance learning (TDL) model has been proposed to effectively make use of space-time information, with a top-push constraint to quantify ambiguous video representation \cite{TDL}.

Deep learning offers an approach to solve feature representation and metric learning problem at the same time. The typical architecture is composed of two parts: a feature extracting network, usually a CNN or RNN, and multiple metric learning layers to make final prediction. The first Siamese-CNN (SCNN) structure \cite{SCNN} proposed for person re-id leveraged a set of 3 SCNNs to the three overlapped parts of the image.
\cite{RFA} exploited a novel recurrent feature aggregation framework, which is capable of learning discriminative sequence level representation for frame-wise features.
A recent work \cite{C_RNN} used CNN to obtain feature representation from the multiple frames of the video, then applied RNN to learn the interaction between them. Temporal pooling layer followed the recurrent layer, aiming to capture sequential interdependence (the pooling might be max-pooling or mean-pooling). Those layers were jointly trained to function as a feature extractor. However, the max-pooling and mean-pooling adapted by them were not robust enough to compress and produce the person's appearance over a period of time, since max-pooling only employed the most active feature map at one temporal step of whole sequence and mean-pooling, which produced a representation averaged over all time steps, thus couldn't preclude the impact of ineffective features well. 

More importantly, the re-id frameworks usually take the form of similarity measure with other inputs. Most of prior works ignored the mutual influence of other items when performing representation learning. Thus we would like to fill this gap by introducing the attention mechanism, which already achieved great success in image caption generation \cite{DBLP:conf/icml/XuBKCCSZB15}, machine translation \cite{DBLP:journals/corr/BahdanauCB14}, question-answering \cite{DBLP:journals/tacl/YinSXZ16} as well as action recognition \cite{DBLP:journals/corr/SharmaKS15}. \cite{DBLP:journals/corr/LiuFQJY16} presented an comparative attention architecture and addressed the problem in the spatial dimension. \cite{Santos2016AttentivePN} proposed a two-way attention mechanism to matching the text sequence, which is exploited in our framework as the temporal pooling component. 

\section{The Proposed Model Architecture}

This work builds a recurrent-convolutional network with jointly attentive spatial-temporal pooling (ASTPN) for video-based person re-identification. Our ASTPN architecture works by passing a pair of video sequences through a Siamese networks to obtain two representations and producing the Euclidean distance between them. As shown in Figure \ref{fig_main}, each input (one frame from a video with optic flow involved) is passed through a CNN network to extract feature maps from the last convolutional layer. Then those feature maps are fed into our spatial pooling layer to obtain image-level representation at one time step. After that, we take temporal information into consideration by utilizing a recurrent network to generate the feature set of a video sequence. Finally, all time steps resulting from recurrent network are combined by attentive temporal pooling to form the sequence-level representation. 

The crucial part of ASTPN relies on the jointly attentive spatial-temporal pooling (ASTP) layers. Instead of using general max (mean) pooling and over-time temporal pooling layers, 
this pooling mechanism could take information  to form the distance at each step, allowing our model to be more attentive both on region of interests in image level and on effective time step in sequence level. Moreover, attentive spatial-temporal pooling also makes our model adaptive to image sequence of arbitrary resolution/length. Detailed techniques about the attentive spatial and temporal pooling will be presented in following subsections.

\subsection{Spatial Pooling Layer}

\begin{figure}[!t]
\centering
\includegraphics[width=8.5cm]{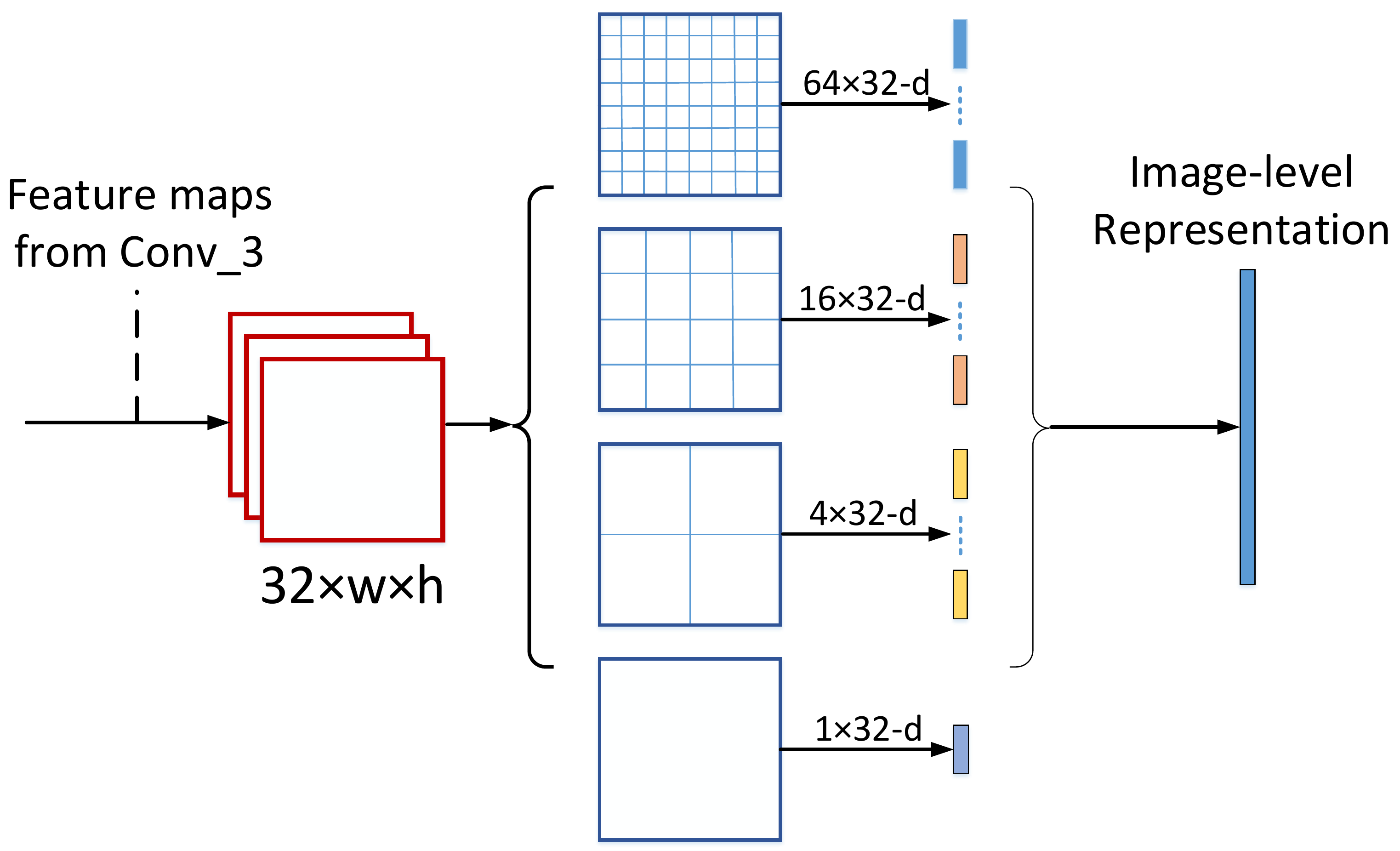}
\caption{Jointly attentive spatial pooling architecture. Here $Conv\_3$ is the last convolutional layer. In the spatial pooling layer, we use a spatial pyramid pooling structure with multi-level spatial bins (8$\times$8, 4$\times$4, 2$\times$2 and 1$\times$1). The image-level representation is then generated by joining all pooling outputs with those spatial bin.}
\label{fig_spatial}
\end{figure}

In person re-identification, due to the overlooking angle of most surveillance equipment, pedestrians only take a part in whole spatial images. Therefore, local spatial attention is necessary for deep networks. The design of such layer should 1) generate multi-scales region patches of each image and feed them into RNN/attention pooling layer; 2) make the model robust to image sequence of arbitrary resolution/length. In this work, we use spatial pyramid pooling (SPP) layer \cite{SPP-net} as the component attentive spatial pooling  to concentrate our model on important region in spatial dimension. Shown in Figure \ref{fig_spatial}, the SPP layer has multi-level spatial bins to generate multi-level spatial representations, and then those representations are combined into a fixed-length image-level representation. Since the image-level representations involve pedestrian position and multi-scale spatial information, our joint attentive spatial pooling mechanism is able to select regions from each frame.

Given the input sequence $\mathbf{v} = \left\{ {{v^1}, \ldots ,{v^T}} \right\}$, we obtain the feature maps set $\mathbf{C} = \left\{ {{C^1}, \ldots ,{C^T}} \right\}$ by utilizing the convolutional network shown in Table \ref{table_structure}. Each ${C^i} \in {\mathbb{R}^{c \times w \times h}}$ is then fed into spatial pooling layer to get image-level representation ${r^i}$. Assuming that the size set of spatial bins is $\left\{ {({m_w}^j,{m_h}^j)|j = 1, \ldots ,n} \right\}$, the window size $wi{n^j} = \left( {\left\lceil {\frac{w}{{m_w^j}}} \right\rceil ,\left\lceil {\frac{h}{{m_h^j}}} \right\rceil } \right)$ and pooling stride $st{r^j} = \left( {\left\lfloor {\frac{w}{{m_w^j}}} \right\rfloor ,\left\lfloor {\frac{w}{{m_w^j}}} \right\rfloor } \right)$ for the $j$-th spatial bin are determined. Then the result vector ${r^i}$ is obtained by formula:
\begin{equation}
\begin{array}{*{20}{c}}
{{b^{i,j}} = {f_R}\left\{ {{f_p}\left( {{C^i};wi{n^j},st{r^j}} \right)} \right\}}\\
{{r^i} = {b^{i,1}} \oplus {b^{i,2}} \oplus  \cdots  \oplus {b^{i,n}}}
\end{array}
\end{equation}
where ${f_p}$ represents the max pooling function with window size $win$ and stride $str$. $\left\lceil  \cdot  \right\rceil $ and $\left\lfloor  \cdot  \right\rfloor $ denote ceiling and floor operations. ${f_R}$ means the reshape operation which reshapes a matrix to a vector. Besides, $ \oplus $ denotes vector connection operation. Let ${\mathbf{r}} = \left\{ {{r^i} \in {\mathbb{R}^L}|i = 1, \ldots ,T} \right\}$ be a sequence representation, where $L = \sum\limits_j {m_w^jm_h^j}$, we then pass $\mathbf{r}$ forward to the recurrent network to extract information between time-steps. The recurrent layer is formulized by:
\begin{equation}
\begin{array}{*{30}{c}}
  {{o^t} = U{r^t} + W{s^{t - 1}},}&{{s^t} = \operatorname{tanh} \left( {{o^t}} \right)} 
\end{array} 
\end{equation}
where ${s^{t - 1}} \in {\mathbb{R}^N}$ is the hidden state containing information from previous time step, and $o^t$ is the output as time $t$. Fully-connected weight $U \in {\mathbb{R}^{L \times N}}$ projects the recurrent layer input $r^t$ from ${{\mathbb{R}^L}}$ to ${{\mathbb{R}^N}}$, and $W \in {\mathbb{R}^{N \times N}}$ projects hidden state ${s^{t - 1}}$ from ${{\mathbb{R}^N}}$ to ${{\mathbb{R}^N}}$. Notice that the recurrent layer embeds the feature vector into a lower-dimensional feature by matrix $U$. The hidden state ${s^{0}}$ is initialized to zero at first time step, and between time steps the hidden state is passed through $tanh$ activation function.

\begin{figure*}[!t]
\centering
\includegraphics[height=6cm]{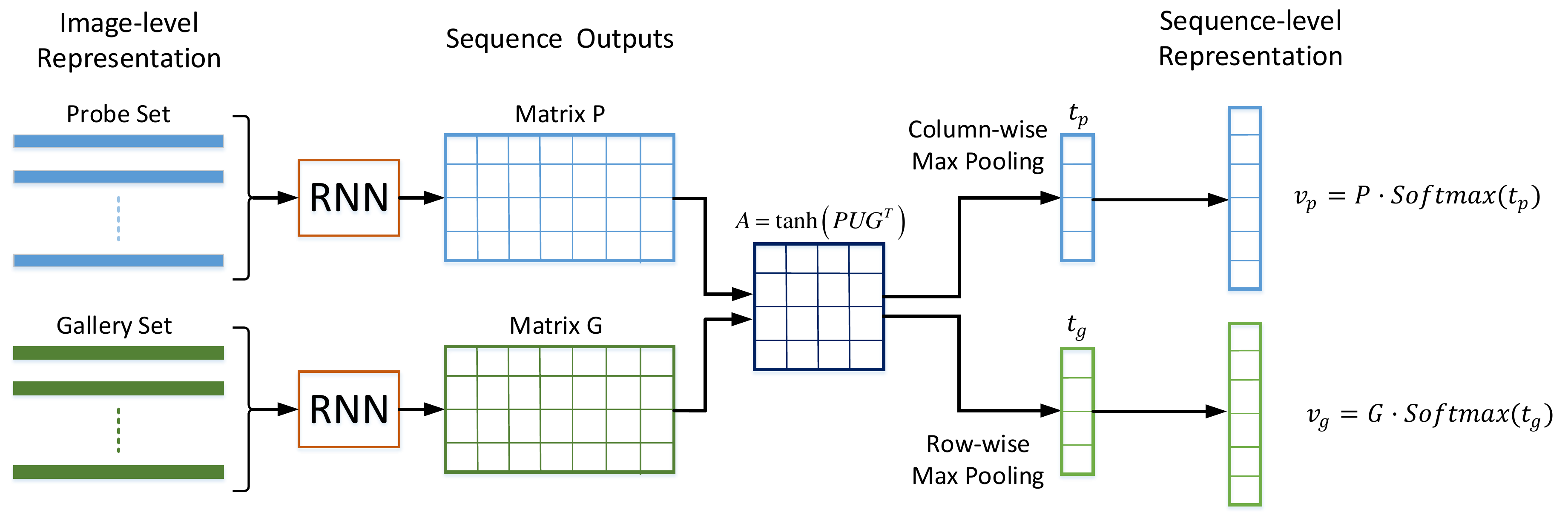}
\caption{Attentive temporal pooling architecture. With the RNNs output matrices $P$ and $G$, we compute attention matrix by introduce a parameter matrix $U$ to capture attentive score in temporal dimension. With column/row-wise max pooling operation and softmax function, the attention vector is obtained, which contains the attentive weight for each time step. The sequence-level representation is computed by dot product between the feature matrices $P$, $G$ and attention vectors $a_p$, $a_g$.}
\label{fig_temporal}
\end{figure*}

\subsection{Attentive Temporal Pooling Layer}

Although the recurrent layer is able to capture temporal information with hidden states, those raw temporarily contains much redundant information. For instance, there are only minor changes in a series of continuous frames, shown in Figure \ref{fig:example}, thus features learned from these sequence input involve a lot of redundant information such as the ambiguous background and clothing. In order to avoid the "Bad money drives out good" issue, we propose an attentive temporal pooling architecture to enable our model to concentrate on effective information. Attentive temporal pooling reinforces the pooling layer to perceive the input probe and gallery data pair, and allows the probe input sequence ${I_P}$ to directly influence the computation of gallery sequence representation $v_g$. We put attentive temporal pooling layer between the recurrent layer and the distance computation layer. In the training phase, attentive temporal pooling is jointly learning with recurrent-convolutional network and the spatial pooling layer, guiding our model for effective information extraction in temporal dimension.

In Figure \ref{fig_temporal}, we illustrate our attentive temporal pooling architecture, which follows the design in \cite{Santos2016AttentivePN,DBLP:journals/tacl/YinSXZ16}. Given the matrices $	P \in {\mathbb{R}^{T \times N}}$ and $G \in {\mathbb{R}^{T \times N}}$, whose $i$-th row represents the output of the recurrent layer in the $i$-th time step with probe data and gallery data respectively, we compute the attention matrix $A \in {\mathbb{R}^{T \times T}}$ as follows:
\begin{equation}
A = \operatorname{tanh} \left( {PU{G^T}} \right)
\end{equation}
where $U \in {\mathbb{R}^{N \times N}}$ is a intent information sharing matrix to be learned by networks. When the convolution and recurrent layer are employed to obtain matrix $P$ and $G$, the attention matrix is able to have a sight on both probe and gallery sequence features, and computes weight scores in temporal dimension. In the gradient descent phase, $U$ is updated by back propagation and influences parameters of convolution and hidden state to guide our model to focus on effective information.  

Next, we apply column-wise and row-wise max pooling on $A$ respectively to obtain temporal weight vector $t_p \in {\mathbb{R}^{T}}$ and $t_g \in {\mathbb{R}^{T}}$. The $i$-th element of $t_p$ represents importance score for the $i$-th frame in the probe sequence, which is the same with $t_g$. Due to the participation of $P$ in the computation of $t_g$, the vector $t_g$ can capture the attentive scores of gallery features related to probe data.

After that, we apply softmax function on temporal weight vectors $t_p$ and $t_g$ to generate attention vectors $a_p \in {\mathbb{R}^{T}}$ and $a_g \in {\mathbb{R}^{T}}$. The softmax function transforms the $i$-th weight ${\left[ {{t_p}} \right]_i}$ and ${\left[ {{t_g}} \right]_i}$ to the attention ratio ${\left[ {{a_p}} \right]_i}$ and ${\left[ {{a_g}} \right]_i}$. For instance, the $i$-th element in $a_g$ is computed as follows:
\begin{equation}
{\left[ {{a_g}} \right]_i} = {{{e^{{{\left[ {{t_g}} \right]}_i}}}} \mathord{\left/
 {\vphantom {{{e^{{{\left[ {{t_g}} \right]}_i}}}} {\sum\limits_{j = 1}^T {{e^{{{\left[ {{t_g}} \right]}_j}}}} }}} \right.
 \kern-\nulldelimiterspace} {\sum\limits_{j = 1}^T {{e^{{{\left[ {{t_g}} \right]}_j}}}} }}
\end{equation}

Finally, we apply dot product between the feature matrices $P$, $G$ and attention vectors $a_p$, $a_g$ to obtain the sequence-level representation $v_p \in {\mathbb{R}^{N}}$ and $v_g \in {\mathbb{R}^{N}}$, respectively:
\begin{equation}
\begin{array}{*{20}{c}}
{{v_p} = {P^T}{a_p},}&{{v_g} = {G^T}{a_g}}
\end{array}
\end{equation}

\subsection{Model Details}

The main thought of our work is to construct a feature extracting network which is able to map the sequence data into feature vector in a low dimensional space, where the feature vectors from the sequences of the same person are close, and the feature vectors from the sequences of different persons are separated by a margin. More details on the components of our proposed network will be explained in follows.

\textbf{Input}: The input to our network consists of three color channels and two optical flow. The color channels provide spatial information such as clothing and background, while optical flow channels provide the temporal motion information. Compared with only use color channels as input, there should be a promotion for person re-id when utilizing both of color channels and optical flow channels .

\textbf{The Siamese Network}: We use a Siamese network architecture as shown in Figure \ref{fig_main}. As mentioned above, our network architecture is grouped by four functional parts: convolutional layers, the attentive spatial pooling layer, the recurrent layer and the attentive temporal pooling layer. As for convolutional layers, we use a convolutional architecture with parameters shown in Table \ref{table_structure}, and the pooling layer in the final layer is replaced by the attentive spatial pooling layer. Notice that convolutional layers are unrolled along with the recurrent layer, and these layers share their parameters in all time steps, which means all frames are passed through the same spatial feature extractor. Similarly, the two recurrent layers also share their parameters to process a pair sequences input. 

\textbf{The Training Objective}: Given a pair of sequences $\left( {{I_p},{I_g}} \right)$ of persons $p$ and $g$, the sequence-level representations $\left( {{v_p},{v_g}} \right)$ are obtained by our Siamese network. After that, we use the Euclidean distance Hinge loss to train our model as follows:
\begin{equation}
E\left( {{v_p},{v_g}} \right) = \left\{ {\begin{array}{*{20}{c}}
  {{{\left\| {{v_p} - {v_g}} \right\|}^2}}&{p = g} \\ 
  {\max \left( {0,m - {{\left\| {{v_p} - {v_g}} \right\|}^2}} \right)}&{p \ne g} 
\end{array}} \right.
\end{equation}
where $m$ denotes the margin to separate features of different persons in Hinge loss. In the training phase, the network is shown positive and negative input pairs alternately. While in the testing phase for a new sequence input, we copy the sequence to form a new pair and pass the pair through our Siamese network to obtain identity feature. By computing the distance between the identity feature with previously saved features of other identities, the most similar identity is indicated with the lowest distance. In addition, we also take identity classification loss into consideration, following the work \cite{C_RNN}. We apply softmax regression on the final features $\left( {{v_p},{v_g}} \right)$ to predict the identity of persons. By using the cross-entropy loss, we obtain the identity loss $I\left( {{v_p}} \right)$ and $I\left( {{v_g}} \right)$. Since that the joint learning of Siamese loss and identity loss brings about a great promotion, the final training objective is the combination of the Siamese loss and the identity loss $L\left( {{v_p},{v_g}} \right) = E\left( {{v_p},{v_g}} \right) + I\left( {{v_p}} \right) + I\left( {{v_g}} \right)$.

\begin{table}[!t]
\newcommand{\tabincell}[2]{\begin{tabular}{@{}#1@{}}#2\end{tabular}}
\renewcommand{\arraystretch}{1.3}
\caption{layer parameter of the CNN network}
\label{table_structure}
\centering
\begin{tabular}{|c|c|c|c|}
\hline
Layer & Type & \tabincell{c}{Conv(size, \\channel, pad, stride)} & Max Pooling\\
\hline
\hline
Conv\_1 & c+t+p & 5$\times$5, 16, 4, 1 & 2$\times$2\\
\hline
Conv\_2 & c+t+p & 5$\times$5, 32, 4, 1 & 2$\times$2\\
\hline
Conv\_3 & c+t & 5$\times$5, 32, 4, 1 & N/A\\
\hline
\end{tabular}
\leftline{}
\leftline{c: Convolutional layer; t: Tanh layer; p: Pooling layer}
\end{table}

\section{Experimental Results}

We evaluate our model for video-based person re-id on three different datasets: iLIDS-VID \cite{video_ranking}, PRID-2011 \cite{PRID} and MARS \cite{MARS}. We also investigate how the joint pooling strategy can bring benefit to the proposed network, the difference between adapting attentive temporal pooling and other common temporal pooling strategies, and the use of attentive spatial pooling.  

\subsection{iLIDS-VID \& PRID-2011}

The iLIDS-VID dataset \cite{video_ranking} contains 300 people in total, where each person is represented by two image sequences recorded by a pair of non-overlapping cameras. The length of frames forming each image sequence ranges from 23 to 192, with an average length of 73.  The challenging dataset was created at an airport arrival hall under a multi-camera CCTV network, whose image sequences were accompanied by clothing similarities among people, lighting and viewpoint variations, cluttered background and occlusions.
 
The PRID-2011 re-id dataset \cite{PRID} consists of 400 image sequences for 200 people captured by two cameras that are adjacent to each other. Each image sequence is composed of frames of length from 5 to 675, with an average number of 100. It's captured in relatively simple environments with rare occlusions, compared with the iLIDS-VID dataset. 
   
\subsubsection{Experiment Settings}
     
Following \cite{C_RNN}, we split the whole set of human sequence pairs of iLIDS-VID and PRID-2011 randomly into two subsets with equal size. One is used for training, and the other is used for testing. We report the performance of the average Cumulative Matching Characteristics (CMC) curves over 10 trials with different train/test splits. Data augmentation was done in several forms. Firstly, since the probe and gallery sequences are of variable-length, sub-sequences of $k = 16$ consecutive frames were chosen randomly at each epoch during training process. Yet we considered the first camera as probe and the second camera as gallery during testing. Secondly, positive pair was composed of a sub-sequence from camera 1 and a sub-sequence from camera 2 containing the same person A, and negative pair was composed of a sub-sequence from camera 1 of person A and a sub-sequence from camera 2 of person B, who was selected arbitrarily from the rest of people in training set.
Positive and negative sequence pairs were sent to our system successively so that the model is capable of distinguishing correct match and wrong match. Lastly, the image level augmentation was performed by cropping and mirroring. Sub-image of both width and length 8 pixels less than its progenitor was produced after cropping, and then we fixed the cropping area within the same sequence. Mirroring operation was randomly applied to a whole sequence together with a probability of $p = 0.5$. Test data also underwent the augmentation to eliminate bias.

Preprocessing steps included the following actions \cite{C_RNN}: Images were converted to YUV color space firstly, and each color channel was normalized to have zero mean and unit variance; Optical flow, both vertical and horizontal, were extracted between each pair of adjoining images using the Lucas-Kanade method \cite{Lucas-Kanade}, and then optical flow channels were normalized to the range -1 to 1; The learning rate, when network was trained with stochastic gradient descent, was 0.001 at the beginning, with batch size set as one. 
   
   The initialization of hyper-parameters of convolutional network was performed based on \cite{C_RNN}, optimized already on the challenging VIPeR person re-identification dataset \cite{parameter_initialization}. Besides, the margin in the Siamese cost function was set to 3, and the dimension of feature space was set to 128. We alternately showed our Siamese network positive and negative sequence pairs, and a full epoch consisted of the equal number of both. As the training set contains 150 people with a maximum sequence length of 192, it takes approximately 3 hours to train for 700 epochs, using the Nvidia GTX-1080 GPU.

 
 \begin{table}[tp]
 \small
 \renewcommand{\arraystretch}{1.3}
 \caption{Comparison of our model with other state-of-the-art methods on iLIDS-VID and PRID-2011 according to CMC curves (\%).}
 \label{table_1}
 \centering
 \begin{tabular}{|c|c|c|c|c|c|c|c|c|}
 \hline
 Dataset & \multicolumn{4}{c|}{iLIDS-VID} &\multicolumn{4}{c|}{PRID-2011}\cr\cline{1-5}
 \hline
 CMC Rank & 1 & 5 & 10 & 20  & 1 & 5 & 10 & 20 \cr
 \hline
 \hline

 ASTPN                &  62  &  86   &  94  &  98   &  77  &  95   &  99  & 99\cr
 \hline
 RNN-CNN \cite{C_RNN}   &  58  &  84  & 91 & 96  & 70 & 90  & 95 & 97\cr
 \hline
  RFA \cite{RFA}         &  49 &  77 & 85   & 92 & 64   &  86  &  93  & 98 \cr
 \hline 
STA \cite{STA}          &  44  &   72  & 84  & 92  & 64   &  87  & 90  &92 \cr
 \hline
  VR \cite{video_ranking} & 35   &  57  & 68 & 78  & 42   & 65   & 78  &89 \cr
  
  \hline
  AFDA \cite{AFDA}      &   38 &   63 &  73  & 82 &  43  &  73   & 85   & 92\cr
 
  \hline
 \end{tabular}

 \end{table}


 \subsubsection{Results}
 We display the results on iLIDS-VID and PRID-2011 in Table \ref{table_1}. The competitor methods are introduced as follows:
 
 \begin{itemize}
 \item RNN-CNN: A recurrent convolutional network (RCN) \cite{C_RNN} with temporal pooling. 
 
 \item RFA: Recurrent feature aggregation network \cite{RFA} based on LSTM, which aggregates the frame-wise human region representation at each time stamp and produces a sequence-level representation.

 \item STA: Spatio-temporal body-action model that takes the video of a walking person as input and builds a spatio-temporal appearance representation for pedestrian re-identification. 
 
 \item VR: A DVR framework presented in \cite{video_ranking} for person re-id uses discriminative space-time feature selection to automatically discover and exploit the most reliable video fragments. 
 

 \item AFDA: An algorithm \cite{AFDA} that hierarchically clusters image sequences and uses the representative data samples to learn a feature subspace maximizing the Fisher criterion.
 
 
 \end{itemize}
 
 Comparing the CMC results of our proposed architecture with the RNN-CNN method and other systems on iLIDS-VID, we can conclude that the attentive mechanism enables our network to outperform all mentioned networks by a large margin. Note that even for the rank-1 matching rate, our method also achieves 62\%, exceeding the RNN-CNN method by more than 4\%. We further notice that even without attentive spatial pooling layer, the utilization of attentive temporal pooling can still lead to a fairly good performance. Thus our proposed network is capable of capturing frame-level human features and then fusing them into a discriminative representation. Another point is that the performances of DNNs seem to apparently surpass the existing state-of-the-art algorithms \cite{SRID,AFDA,DTDL}, proving the power of DNNs when sufficient training data is available.
 
Less challenging than iLIDS-VID as PRID-2011 is, we can observe overall increments in matching rate. Our model still outperforms other methods prominently in terms of Table \ref{table_1}, with rank-1 accuracy achieving 77\%---transcending  the RNN-CNN method by 7\%. Besides, our system is more efficient and robust since its CMC rank rate reaches 95\% at level of rank 5 and  further goes up to the summit of 99\% quickly at the level of rank 10. The tendency of accuracy demonstrates that our system is an effective space-time feature extractor, able to obtain more discriminative sequence-level representation through learning process. DNNs \cite{C_RNN,RFA} still exhibit distinctive capability of capturing human features on the whole, with accuracy converging at earlier point.
 

\begin{figure}[!htb]
\centering
\subfigure[]{
\label{fig:iLIDS-VID}
\includegraphics[height=4.8cm,width=8cm]{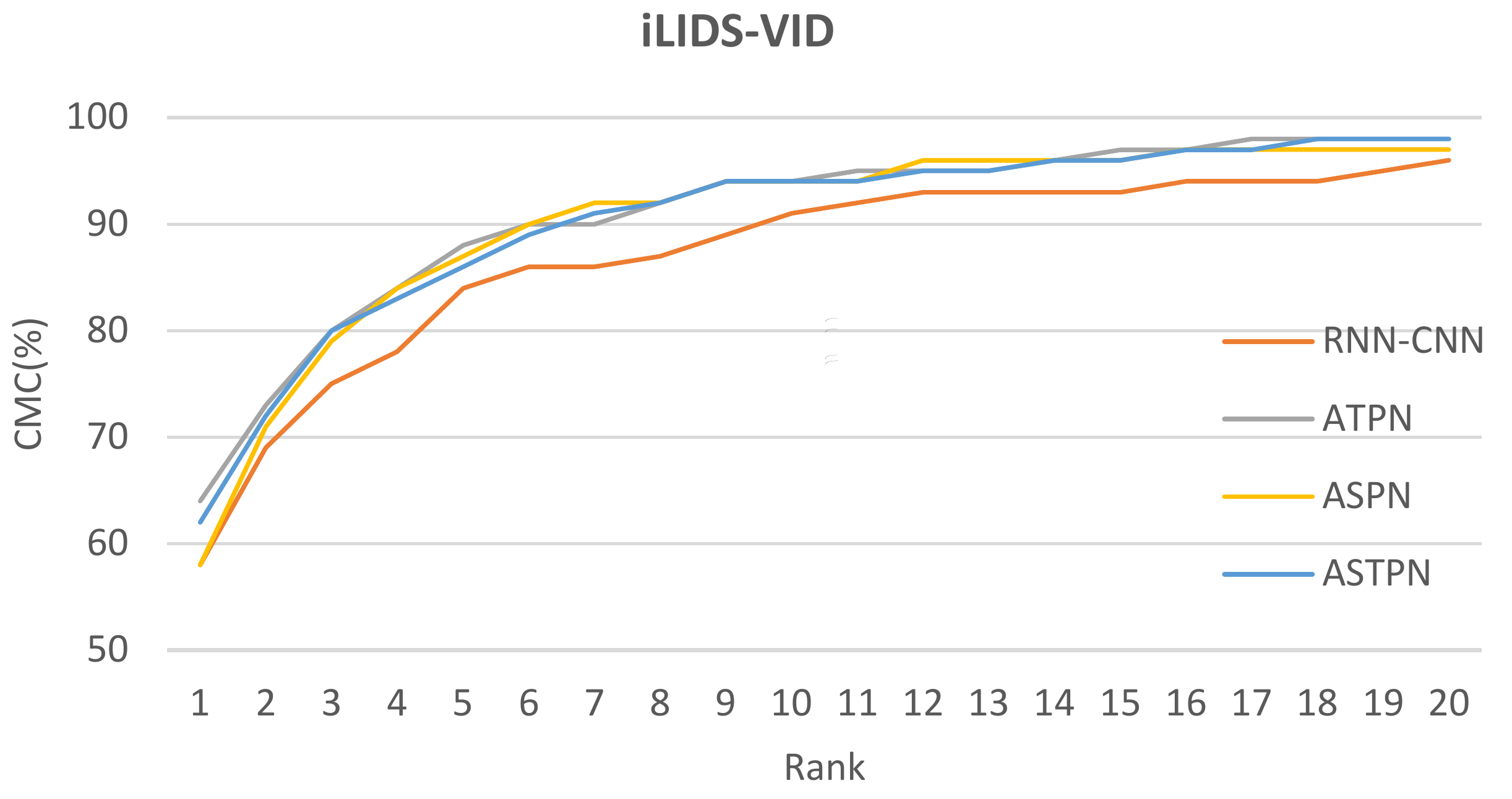}
}
\subfigure[]{
\label{fig:PRID-2011}
\includegraphics[height=4.8cm,width=8cm]{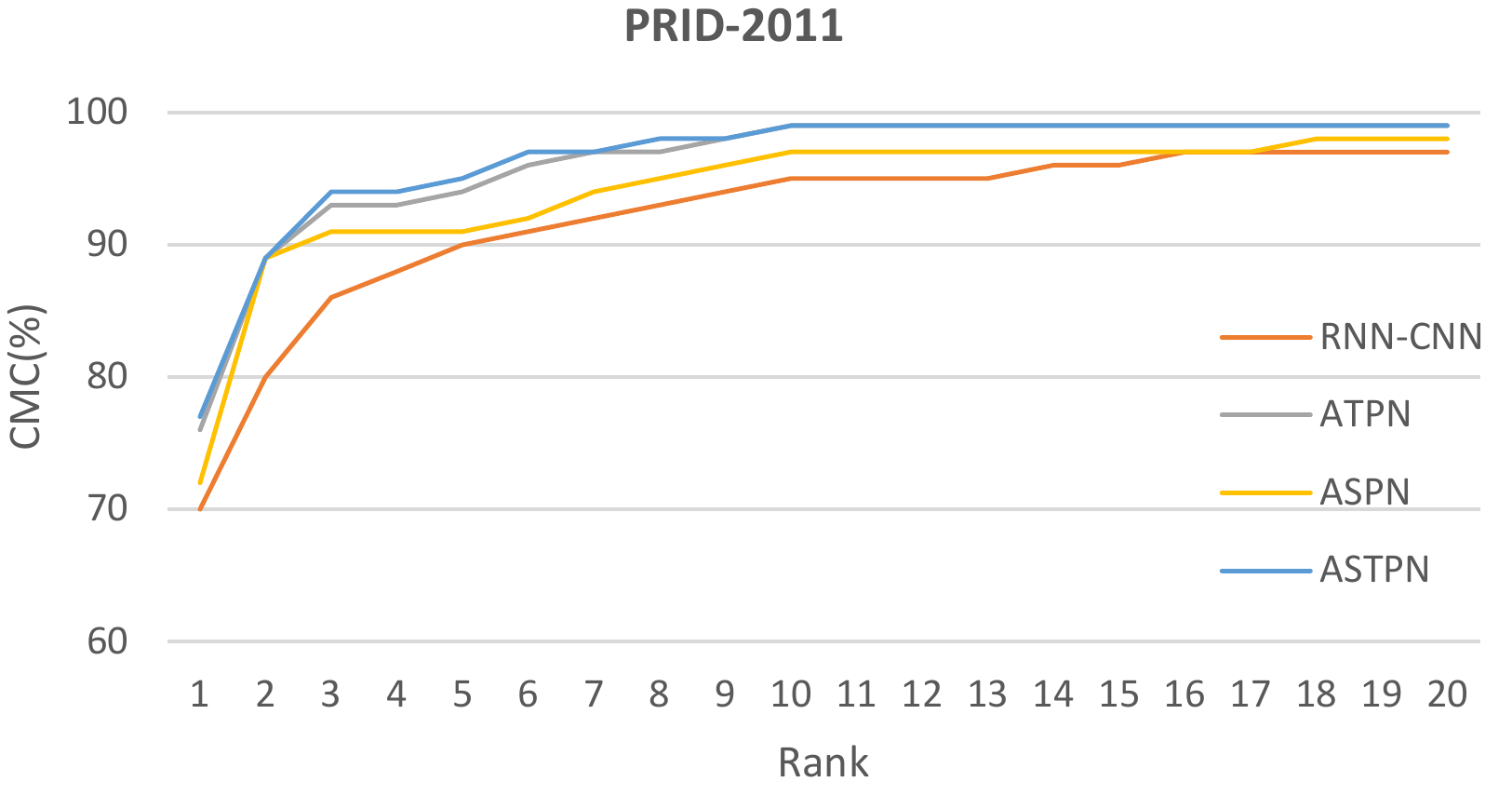}
}
\subfigure[]{
\label{fig:MARS}
\includegraphics[height=4.8cm,width=8cm]{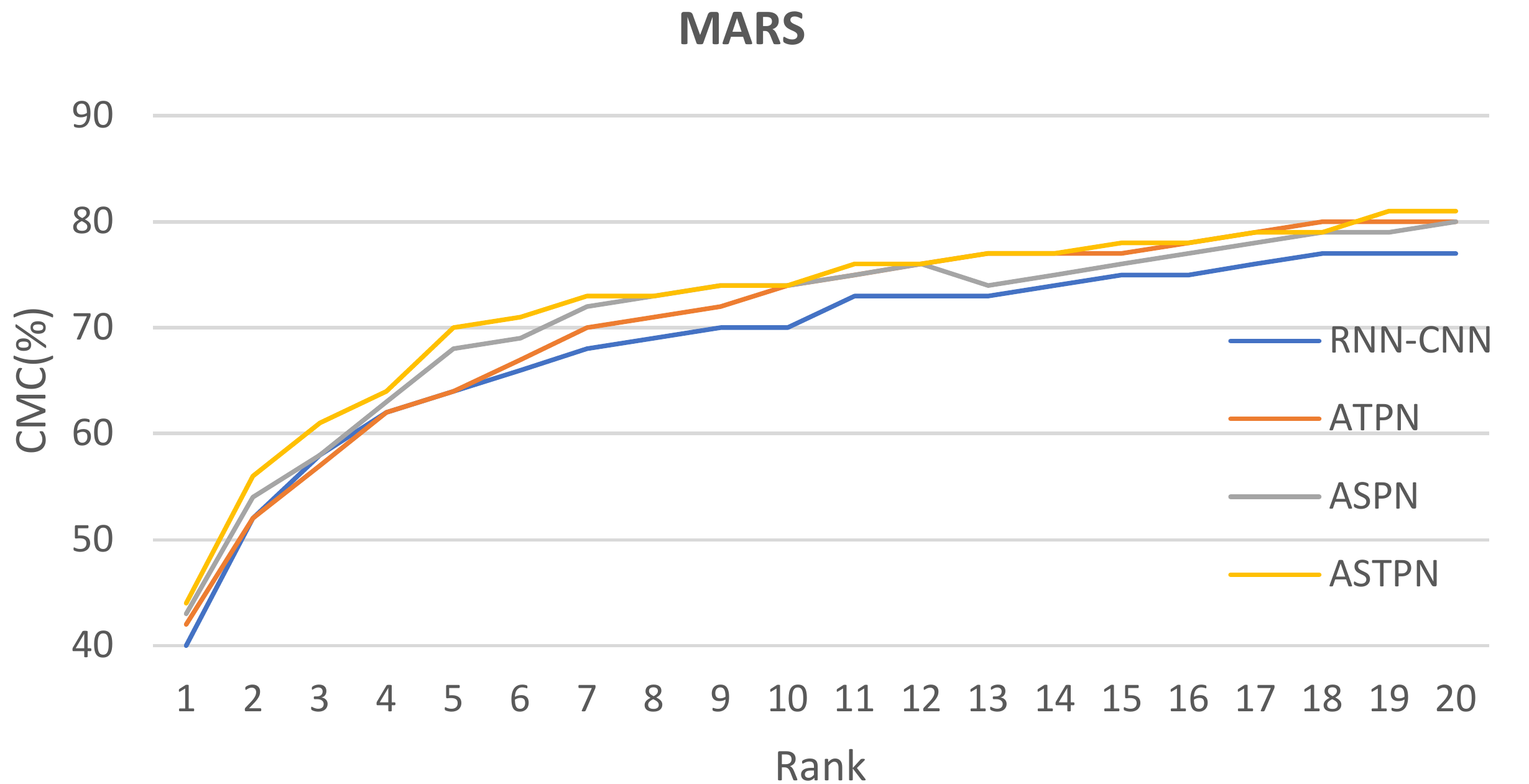}
}
\caption{The variants of our model are tested on three datasets respectively. ATPN refers to attentive temporal pooling network and ASP refers to attentive spatial pooling network. Finally ASTPN stands for the combination of ATPN and ASPN.}
\label{fig:Comparison}
\vspace{-.8em}
\end{figure}

\subsection{MARS}
This is a dataset introduced in \cite{MARS}, which is also claimed to be the largest video re-id dataset to date. MARS consists of 1261 different pedestrians, each of whom was captured by at least two cameras.
Compared with iLIDS-VID and PRID-2011, MARS is 4 times larger in the number of identities and 30 times larger in total tracklets. The tracklets of MARS are generated automatically by DPM detector and GMMCP tracker, whose error makes MARS more realistic and of course more challenging than previous dataset. Each identity has 13.2 tracklets on average. For instance, most identities are captured by 2-4 cameras, and most identities have 5-15 tracklets, most of which contain 25-50 frames.

To perform our experiments on MARS, simplification should be done in two steps. Firstly, as pedestrians were recorded by at least 2 cameras, we randomly chose 2 camera viewpoints of the same person out of the ensemble. Then one of them was set as probe set and the other was set as gallery set. Here the case was reduced to our previous experiences with iLIDS-VID and PRID-2011. 
 
The performances of our models are displayed in Table \ref{table_MARS}, compared with the baseline RNN-CNN. 
ASTPN still achieves the best accuracy while general results dropping obviously in contrast with Table \ref{table_1}. Compared with iLIDS-VID and PRID-2011, the improvement is larger (around 4\% in all ranks). The reasons may be attributed to that a considerable part of image sequences of MARS are accompanied by cluttered backgrounds, ambiguity in visual appearance, or drastic viewpoint changes between sequence pairs.
 
 \begin{table}[tp]
 \small
 \renewcommand{\arraystretch}{1.3}
 \caption{Performance comparison with CMC Rank accuracy on MARS (\%). }
 \label{table_MARS}
 \centering
 \begin{tabular}{|c|c|c|c|c|}
 \hline
 Dataset &\multicolumn{4}{c|}{MARS}\\ \cline{1-5} 
 
 CMC Rank& 1 & 5 & 10 & 20\\
 \hline
 \hline
 RNN-CNN &       40  & 64   &  70 &  77     \\
 \hline
 ASTPN &         44   &  70  & 74   &   81    \\
 \hline
 
 \end{tabular}

 \end{table}


  


\subsection{Control Experiments with Different Pooling Strategies}

We investigate the effects of attentive temporal pooling (ATPN), attentive spatial pooling (ASPN), and the coexistence of them (ASTPN). The related CMC curves on iLIDS-VID, PRID-2011 and MARS are presented in Figure \ref{fig:Comparison} respectively. 

\textbf{ATPN}: the overall performance of ATPN curve is obviously better than the RNN-CNN method in Figure \ref{fig:iLIDS-VID} and \ref{fig:PRID-2011}. For example, ATPN curve exceeds the RNN-CNN method by almost 10\% at the rank 2 accuracy on PRID-2011. Meanwhile, on iLIDS-VID ATPN curve also outperforms the RNN-CNN method by 5\% at the rank 3 accuracy. We can safely conclude that ATPN can efficiently utilize temporal human appearance to form powerful sequence-level representation, which is more subtle and discriminative than the output of simple pooling strategies (max-pooling and mean-pooling).       

\textbf{ASPN} exhibits equally prominent capability of matching compared with ATPN on iLIDS-VID. Although it is less robust than ATPN on PRID-2011, distinct margin still exists between ASPN curve and the RNN-CNN method. We may reason that ASPN, attentive spatial pooling network, mainly leverages relevant contextual information to enhance the discriminative power of the final representation. However, as we have mentioned about these two datasets, iLIDS-VID was created in a rather complicated environment, which means the contextual information could be more valuable clue due to the ambiguity of human appearance. On the contrary, ASPN thus doesn't perform as competitively as ATPN in Figure \ref{fig:PRID-2011}.    

\textbf{ASTPN}: combining ATPN and ASPN together, ASTPN can capitalize on frame-wise interactions effectively as well as selectively propagate additional contextual information through the network. Based on Figure \ref{fig:MARS}, where apparent distinction between ATPN curve and ASTPN curve can be observed with overall accuracy decreasing caused by MARS, ASTPN exceeds ATPN by about 5\% at rank 3 point. It's proven that ASTPN is a more robust joint method especially on dataset as challenging as MARS. 
 
\subsection{Cross-Dataset Testing}
Data bias is inevitable since a particular dataset only represents a small fragment of data of whole real world. The machine-learning model trained on A dataset would perform much worse when tested on B dataset. It can be regarded as over-fitting to the particular scenario, thus reducing the generality of the model. Cross-data testing is designed to evaluate the model's potentials in practical application.

The settings are introduced as follows: Both ASTPN and RNN-CNN are trained on diverse iLIDS-VID dataset, and then are tested on 50\% of the PRID-2011 dataset. Apart from distinction brought out by cross-dataset training, the contrast of single-shot method and multi-shot method is also shown in Table \ref{table_Cross}. Although results are much worse than Table \ref{table_1}, ASTPN still achieves 30\% on rank 1 accuracy, close to  SRID \cite{SRID} which is trained on PRID-2011 with the rank 1 accuracy of 35\% . Moreover, using video-based re-id seems to improve the scores of both models by 100\% than using single-shot re-id in terms of rank 1 score. It can be concluded that the valuable temporal information provided by video-based re-id really enhance the generalization performance of re-id networks greatly.

 
 \begin{table}[tp]
 \small
 \renewcommand{\arraystretch}{1.3}
 \caption{Cross-dataset testing matching rate on PRID-2011 (\%). * indicates that both probe set and gallery set are composed of single image during test.}
 \label{table_Cross}
 \centering
 \begin{tabular}{|c|c|c|c|c|c|}
 \hline
 Model & Trained on & 1 & 5 & 10 & 20\\
 \hline
 \hline
 ASTPN &      iLIDS-VID      &  30  & 58  &  71  &  85  \\
 \hline
 ASTPN* &       iLIDS-VID     &  15  & 33  &  46  &  63  \\
 \hline
 RNN-CNN &    iLIDS-VID    &    28 &  57  &  69 &  81     \\
 \hline
 RNN-CNN* &     iLIDS-VID    &   14 &  31  & 45  &  61     \\
 \hline
 
 \end{tabular}

 \end{table}


\section{Conclusion}
We proposed ASTPN, a novel deep architecture with jointly attentive spatial-temporal pooling for video-based person re-identification, enabling a joint learning of the representations of the inputs as well
as their similarity measurement. ASTPN extends the standard RNN-CNNs by decomposing pooling into two steps: a spatial-pooling on feature map from CNN and an attentive temporal-pooling on the output of RNN. In effect, explicit or implicitly attention is performed at each pooling stage which 
select key regions or frames over the sequences for the feature representation learning. 

Extensive experiments on iLIDS-VID, PRID-2011 and MARS have demonstrated that ASTPN significantly outperforms standard max and temporal pooling approaches. In particular, by executing control experiments, we show the joint pooling power than either of spatial/temporl pooling separately. Additionally, ASTPN is simple to implement and introduces little computational overhead compared to general max pooling, which makes
it a desirable design choice for deep RNN-CNNs used in person re-identification in future. We would also consider to apply the current method into target tracking/detection systems \cite{Wang_2016_CVPR}. 

\section{Acknowledgment}
This research is supported by NSFC No. 61401169. The corresponding author is Pan Zhou.

{\small
\bibliographystyle{ieee}
\bibliography{egbib}
}

\end{document}